\documentclass[acmtog,screen]{acmart}

\AtBeginDocument{%
  }

\setcopyright{acmlicensed}
\copyrightyear{2025}
\acmYear{2025}
\acmDOI{3757377.3763948}
\acmConference[SA Conference Papers '25]{SIGGRAPH Asia 2025 Conference Papers}{December 15--18,
  2025}{Hong Kong}

\usepackage{booktabs} % For formal tables

\usepackage{xcolor,colortbl}
\usepackage{color}

\usepackage[ruled]{algorithm2e} % For algorithms

\usepackage{url}
\usepackage[bb=boondox]{mathalfa}
\usepackage{multirow}

\SetAlFnt{\small}
\SetAlCapFnt{\small}
\SetAlCapNameFnt{\small}
\SetAlCapHSkip{0pt}

\usepackage{multirow}
\usepackage{graphicx}

\usepackage{xspace}
\usepackage{subfig}
\providecommand{\FullStop}{\text{~\@.\xspace}}
\providecommand{\Comma}{\text{~,\xspace}}

\begin{document}
\title{Learning Human Motion with Temporally Conditional Mamba}

\author{Quang Nguyen}
\orcid{0009-0007-1831-5763}
% \authornotemark[1]
\affiliation{%
  \institution{FPT Software AI Center}
  \city{Hanoi}
  \country{Vietnam}
}
\email{quangnv89@fpt.com}

\author{Tri Le}
\orcid{0000-0002-2546-5651}
\affiliation{%
  \institution{FPT Software AI Center}
  \city{Hanoi}
  \country{Vietnam}}
\email{trilq3@fpt.com}

\author{Baoru Huang}
\orcid{0000-0002-4421-652X}
\affiliation{%
  \institution{University of Liverpool}
  \city{Liverpool}
  \country{United Kingdom}
}
\email{baoru.huang@liverpool.ac.uk}
\authornote{Corresponding author.}

\author{Minh Nhat Vu}
\orcid{0000-0003-0692-8830}
\affiliation{%
 \institution{Vienna University of Technology}
 \city{Wien}
 \country{Austria}}
\email{minh.vu@tuwien.ac.at}

\author{Ngan Le}
\orcid{0000-0003-2571-0511}
\affiliation{%
  \institution{University of Arkansas}
  \city{Arkansas}
  \country{USA}}
\email{thile@uark.edu}

\author{Thieu Vo}
\orcid{0000-0001-7957-5648}
\affiliation{%
  \institution{National University of Singapore}
  \city{Singapore}
  \country{Singapore}}
\email{thieuvo@nus.edu.sg}

\author{Anh Nguyen}
\orcid{0000-0002-1449-211X}
\affiliation{%
  \institution{University of Liverpool}
  \city{Liverpool}
  \country{United Kingdom}}
\email{anh.nguyen@liverpool.ac.uk}

\renewcommand{\shortauthors}{Quang Nguyen et al.}

\begin{abstract}
Learning human motion based on a time-dependent input signal presents a challenging yet impactful task with various applications. The goal of this task is to generate or estimate human movement that consistently reflects the temporal patterns of conditioning inputs. Existing methods typically rely on cross-attention mechanisms to fuse the condition with motion. However, this approach primarily captures global interactions and struggles to maintain step-by-step temporal alignment. To address this limitation, we introduce Temporally Conditional Mamba, a new mamba-based model for human motion generation. Our approach integrates conditional information into the recurrent dynamics of the Mamba block, enabling better temporally aligned motion. To validate the effectiveness of our method, we evaluate it on a variety of human motion tasks. Extensive experiments demonstrate that our model significantly improves temporal alignment, motion realism, and condition consistency over state-of-the-art approaches. Our project page is available at \href{https://zquang2202.github.io/TCM}{https://zquang2202.github.io/TCM}.
\end{abstract}




\maketitle

\section{Introduction}
\label{sec:intro}
Learning human motion has received significant interest from multiple research communities, including computer vision, computer graphics, machine learning, and multimedia~\cite{gao2022pc, kim2022brand, guo2022generating}. This area typically involves two main tasks: human motion synthesis and human motion estimation, both of which support various applications such as animation, virtual reality, and human-robot interaction~\cite{nishimura2020long, zhu2023human}. The goal of human motion synthesis is to generate realistic and natural human motion conditioned on the inputs, such as audio~\cite{li2021ai, tseng2023edge, li2024lodge}, text~\cite{tevet2023human, pinyoanuntapong2024bamm, zhang2024motion}, environmental context~\cite{huang2023diffusion, li2023object}, or video~\cite{li2023ego, wang2023scene, hong2024egolm}. In contrast, human motion estimation aims to predict human motion from observed input signals. A core challenge in human motion synthesis and estimation tasks is effectively capturing the complex dependencies between motion dynamics and external stimuli. For example, in music-driven motion generation, the synthesized human motion should be physically plausible and rhythmically aligned with the music's beat.

\begin{figure*}[ht]
    \centering
    \includegraphics[width=1\linewidth]{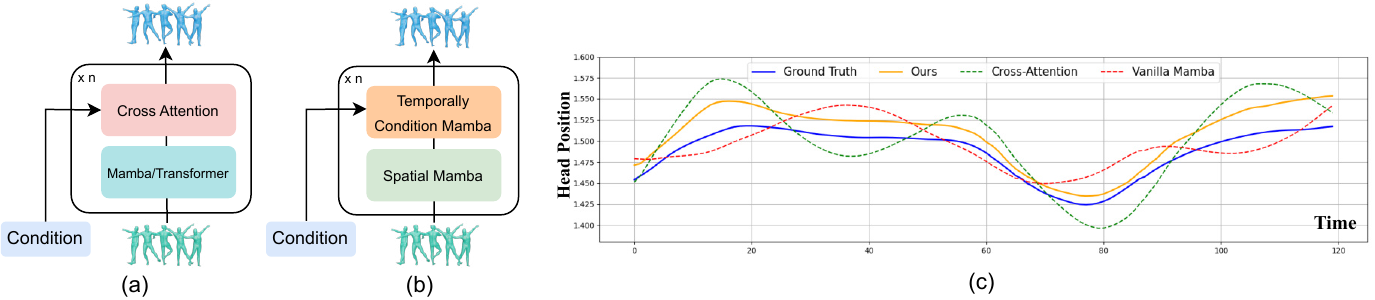}
    \caption{High-level comparison between our approach and previous methods. (a) Previous works usually use Cross-Attention to integrate input condition into Mamba/Transformer backbone. (b) Our approach embeds the condition directly within the Mamba block; (c) We show the head trajectory over time in an ego-to-motion task. Compared to Cross-Attention and Vanilla Mamba, which generate motions that deviate noticeably from the ground truth, our method produces a trajectory that closely follows the actual motion pattern.}
    \vspace{-1ex}
    \label{fig:intro}
\end{figure*}

Several works on human motion have recently focused on \textit{static} conditional inputs such as text description in the text-to-human motion task~\cite{tevet2023human, yuan2024mogents,zhang2024motiondiffuse,petrovich2022temos}. The static condition input provides a high-level semantic intent or partial temporal and spatial constraints, but remains constant throughout the generated motion sequence. In contrast, another line of work addresses \textit{temporal} conditioning inputs such as music~\cite{chan2019everybody, tseng2023edge}, egocentric video~\cite{li2023ego, wang2024egocentric, luo2021dynamics}, time-series signal~\cite{huang2024beat,lam2022human, tang2024unified}, or tracking inputs~\cite{starke2022deepphase, du2023avatars, starke2024categorical}. These temporal conditions evolve and carry rich, fine-grained temporal dynamics that directly influence the temporal dynamics of the human motion. This setting poses a greater challenge, requiring the model to continuously adapt to changing inputs while maintaining physical plausibility and temporal coherence. In this work, we explore human motion problems under the \textit{temporal conditions}, aiming to capture precise alignment between human motion and the evolving temporally conditional input.

%Fig.~\ref{fig:intro} (c) illustrates the mean joint velocity over time for a representative dance sequence. We observe that motion beats, identified as local minima in kinetic velocity, produced by our method, exhibit stronger temporal alignment with music beats (highlighted by orange dashed rectangle in Fig.~\ref{fig:intro} (c)) compared to baselines. 

To synthesize or estimate human motion from temporal conditions, previous works have mainly utilized the Cross-Attention mechanism within diffusion framework~\cite{ho2020denoising}. These frameworks typically use either Mamba-based~\cite{zhang2024motion, hu2024zigma, gu2023mamba} or Transformer-based~\cite{tseng2023edge, li2024lodge} architectures as their backbone (Fig.~\ref{fig:intro}a). For example, EgoEgo~\cite{li2023ego} generates human motion from egocentric views using a transformer decoder network. Other works~\cite{tseng2023edge, li2024lodge, huang2024beat} use Cross-Attention to align music with dance motion. While notable results have been achieved with handling temporal conditions using the Cross-Attention mechanism, this approach often overlooks recurrent dependencies and fine-grained temporal alignment. %Furthermore, cross-attention captures global context but neglects temporal structure. 
In Fig.~\ref{fig:intro}c, we illustrate the head position over time for a generated sequence in the ego-to-motion task. We can see that both Cross-Attention and Vanilla Mamba fail to produce trajectories that closely match the ground truth%, while our method generates a trajectory with significantly higher similarity.
. Therefore, we hypothesize that the conditioning signal should be modeled as a recurrent influence on the motion stream to better preserve temporal structure and alignment of human motion. To this end, we propose a new method that directly integrates the condition into Mamba’s dynamics, enabling the autoregressive injection of temporal signals to improve the coherence and alignment between human motion and condition inputs.

% \vspace{-2ex}
Building on the widely adopted diffusion framework known for its effectiveness in generating high-quality motion, we introduce a new approach for human motion learning. Specifically, we propose Temporally Conditional Mamba (\textbf{TCM}), a new Mamba block that can be integrated into the diffusion model for human motion tasks. Our method integrates temporal conditions into each human motion timestep, allowing the model to learn human motion that is temporally aligned with the input conditions. We demonstrate TCM's generalizability across diverse tasks in both human motion \textit{synthesis} or human motion \textit{estimation} settings with different temporally conditional inputs. Extensive experiments show that our approach consistently outperforms state-of-the-art methods in motion quality and condition alignment. Our main contributions can be summarized as follows:
\begin{itemize}
    \item We introduce TCM, a new mechanism that injects temporal conditions into the Mamba's dynamics, enabling autoregressive alignment of human motion with condition signals.
    \item We show that TCM can be integrated into a diffusion model for diverse motion synthesis and motion estimation tasks with temporal conditions, and achieves significant improvements over state-of-the-art methods.
    % across a variety of human motion-related tasks, ranging from motion synincluding dance synthesis from music, motion estimation from egocentric video, motion estimation from egocentric and music, and motion prediction from object trajectories.
    %\item We empirically demonstrate that our approach achieves significant improvements over state-of-the-art methods in terms of motion-condition alignment, motion realism, and physical plausibility across multiple tasks.
\end{itemize}

% \vspace{-4ex}
\section{Related Works}
\label{sec:relatedwork}
\subsection{Human Motion.} Human motion is a popular research topic~\cite{rempe2021humor, petrovich2021action, jiang2023motiongpt, zhang2023generating,chen2023executing}. Several works have focused on human motion synthesis with static conditions, such as text-to-human motion~\cite{zhang2024motiondiffuse,dai2024motionlcm,petrovich2022temos,tevet2023human}. In practice, human motion can be synthesised or estimated based on temporal conditions such as music~\cite{chan2019everybody,siyao2022bailando, lam2022human, tseng2023edge, le2023music, le2023controllable}, object trajectories~\cite{taheri2022goal, li2023object}, videos~\cite{li2023ego,mehraban2024motionagformer,zhao2024egobody3m, chan2019everybody},  speech~\cite{chhatre2024emotional}, and multimodal combinations of egocentric video and music~\cite{nguyen2025egomusic}. 
% How current papers condition the motion synthesis
Current works leverage simple concatenation or cross-attention~\cite{vaswani2017attention} to condition human motion on input modalities. The authors in~\cite{luo2021dynamics, yuan2019ego} concatenate optical flow features and pose states as input to a policy network. The works in~\cite{li2023object, li2023ego} employ a diffusion model where the noisy data is concatenated with object movement trajectories and head pose. %In~\cite{siyao2022bailando}, a method is proposed to place music frame tokens at the beginning of a GPT-like decoder, where the generation of human motion tokens can attend to music tokens. 
The authors in~\cite{tseng2023edge, huang2024beat, li2024lodge} leverage the cross-attention mechanism to inject music information into dance generation at each decoder layer and denoising step. Unlike previous works using the cross-attention mechanism or simple concatenation, our work injects the temporal condition input directly into each decoder layer of the network to enable better alignment between the temporal condition input and the human motion.

\subsection{State Space Models.}
State space models (SSM)~\cite{gu2022efficiently,gu2021combining} are a promising architecture for sequence modeling~\cite{hu2024zigma}. Mamba~\cite{gu2023mamba} introduces a selective SSM architecture, integrating time-varying parameters into the SSM framework. Several works apply Mamba in various applications, including image processing~\cite{hu2024zigma, hatamizadeh2024mambavision, lee2024meteor, phung2024dimsum, wang2024mamba}, graph processing~\cite{behrouz2024graph, wang2024graph}, point cloud analysis~\cite{liang2024pointmamba,liu2024point}, and human motion generation~\cite{zhang2024motion, wang2024text}. Vim~\cite{liao2024vision} and DiM~\cite{teng2024dim} present different scanning strategies for the SSM block in vision tasks. Mamba-ND~\cite{li2024mamba} extends the capabilities of SSM to higher-dimensional data by exploring different scan directions within a single SSM block. ZigMa~\cite{hu2024zigma} and AiM~\cite{li2024scalable} utilize Mamba-based SSM blocks for efficient image generation. The authors in~\cite{yan2024diffusion} propose a fully Mamba-based model that eliminates attention mechanisms, demonstrating its effectiveness for high-resolution image generation. MotionMamba~\cite{zhang2024motion} proposes a symmetric multi-branch Mamba that processes temporal and spatial and performs exceptionally on text-to-motion generation tasks. To enable conditional generation, current Mamba-based architectures~\cite{hu2024zigma, li2024scalable, zhang2024motion} adopt the condition mechanism from their origin Transformer-based ones. In contrast, our method exploits conditioned temporal data in the Mamba block. 
% We may need to say some theory that we rely on because it is our significant difference.

\subsection{Feature Modulation.} Feature-wise modulation learns a function of conditioning inputs to replace parameters in feature-wise affine transformation, as introduced originally in~\cite{ioffe2015batch}. 
Feature-wise modulation has various forms applied across domains: Conditional Instance Normalization~\cite{dumoulin2016learned} and Adaptive Instance Normalization~\cite{huang2017arbitrary} that are applied in image stylization; Dynamic Layer Normalization~\cite{kim2017dynamic}, which is successfully applied for speech recognition. Recently, FiLM~\cite{perez2018film} and Adaptive Normalization~\cite{xu2019understanding} are widely used in image generation~\cite{karras2019style,peebles2023scalable,dhariwal2021diffusion,chen2023pixart}, audio generation~\cite{liu2023audioldm}, and vision language action model~\cite{brohan2022rt, chi2023diffusion,team2024octo,turkoglu2022film}, motion generation~\cite{yan2024diffusion, li2024lodge}. Inspired by the success of feature-wise modulation, we integrate an adaptive layer normalization to modulate the features directly in each Mamba step during the human motion generation process.

\begin{figure*}[t]
    \centering
    \includegraphics[width=\textwidth]{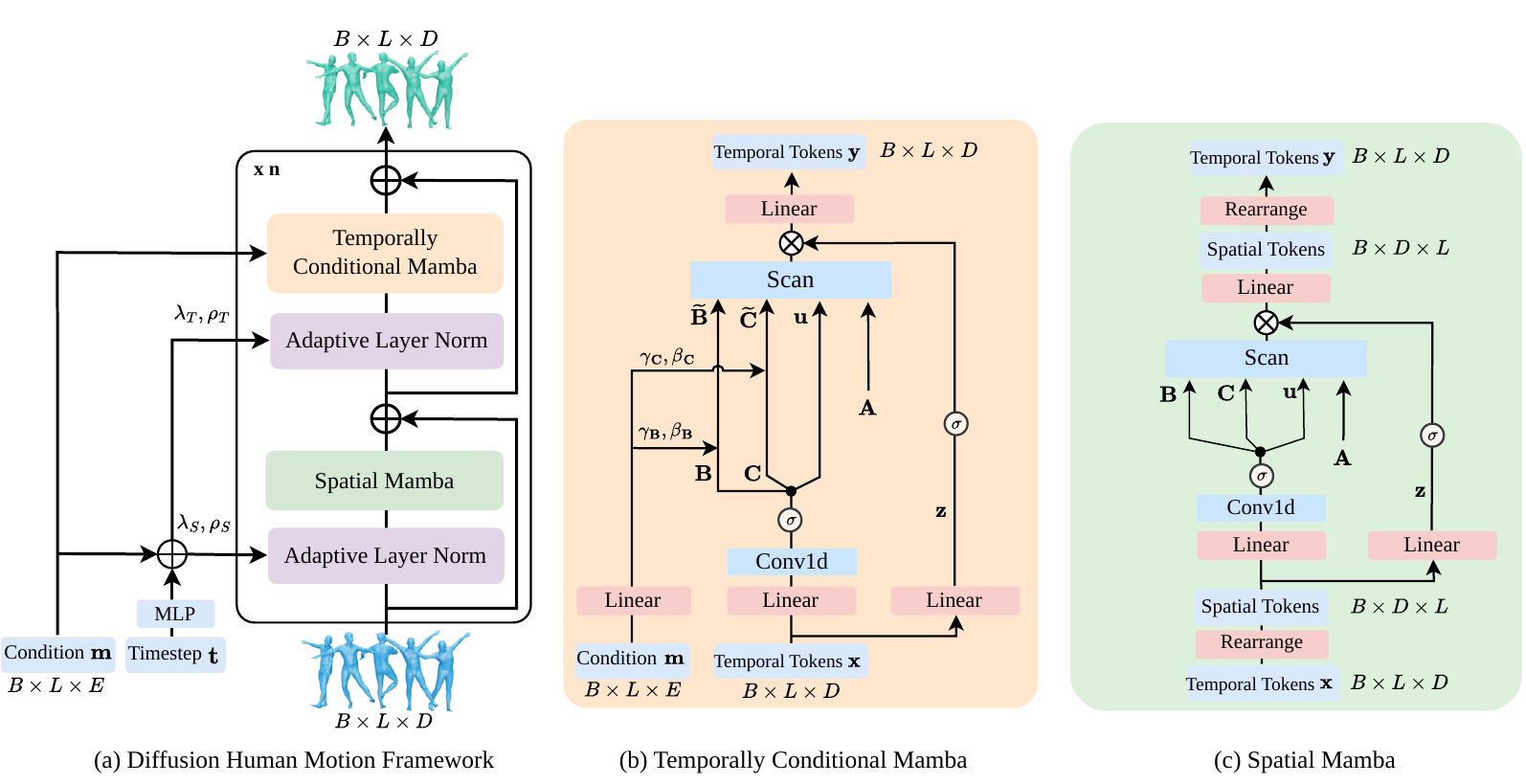}
    \vspace{-4ex}    
    \caption{Architecture overview of the proposed approach. (a) We show the overview of the diffusion human motion framework with Mamba blocks. (b) Our key contribution is the Temporally Conditional Mamba, which incorporates temporal conditions into the internal dynamics of the Mamba block. (c) The Spatial Mamba block is used to learn human spatial features.}
    \vspace{-1ex}
    \label{fig: method}
\end{figure*}
\section{Methodology}

\subsection{Preliminaries}
\subsubsection{Mamba.} A Mamba block~\citep{gu2023mamba} integrates S6 layers as its central element. % alongside components like linear transformation, 1d convolution, and other token-wise operators. The S6 layer, acting as the heart of the Mamba block, is crucial to the effectiveness of these model
An S6 layer operates by transforming an input sequence of tokens \(\mathbf{x} = (x_1, \ldots, x_L)  \in \mathbb{R}^{L\times D} \) to the output sequence of tokens \( \mathbf{y} = (y_1, \ldots, y_L)\in \mathbb{R}^{L\times D} \), where $L$ is the sequence length. In this context, each vector $x_l \in \mathbb{R}^D$ or $y_l\in\mathbb{R}^D$ represents a token in the sequence. For each feature channel indexed by $d \in {1, \ldots, D}$, the output sequence $y_d = (y_{d1}, \ldots, y_{dL}) \in \mathbb{R}^L$ is computed recursively from the corresponding input sequence $x_d = (x_{d1}, \ldots, x_{dL}) \in \mathbb{R}^L$. This computation progresses through a series of hidden states $h_{d1}, \ldots, h_{dL} \in \mathbb{R}^N$, defined as:
\begin{equation}\label{eq: origin_mamba}
    \begin{aligned} 
        h_{dl} &= \overline{\mathbf{A}}_{dl} \cdot h_{d,l-1} + \overline{\mathbf{B}}_{dl} \cdot x_{dl}, \quad h_{d0} = 0\Comma \\ 
        y_{dl} &= \mathbf{C}_l \cdot h_{dl}\Comma
    \end{aligned}
\end{equation}
where, at each timestep \(l = 1, \ldots, L\), the dynamic system matrices \((\overline{\mathbf{A}}_{dl}, \overline{\mathbf{B}}_{l})\) are derived from continuous-time matrices\((\mathbf{A}_{d}, \mathbf{B}_{l})\) through discretization using a time scaling factor $\Delta_{dl}$:
\begin{equation}
    \overline{\mathbf{A}}_{dl} = e^{\Delta_{dl} \cdot \mathbf{A}_d}, \quad
\overline{\mathbf{B}}_{dl} = \Delta_{dl} \cdot\mathbf{B}_{l}\Comma
\end{equation}
where \(\Delta_{dl}(x_l)\), \(\mathbf{B}_l(x_l)\), and \(\mathbf{C}_l(x_l)\) are dynamically computed from the input $x_l$:
\begin{equation}
\begin{aligned}
    &\Delta_{dl}(x_l) = \text{softplus}(S_{\Delta,d}(x_l))=ln(1+e^{S_{\Delta,d}(x_l)}), \\ &\mathbf{B}_{l}=S_{\mathbf{B}}(x_l),\quad \mathbf{C}_l= (S_{\mathbf{C}}(x_l))^\top \FullStop
\end{aligned}
\end{equation}
The hidden matrix \(\mathbf{A}_d \in \mathbb{R}^{N\times N}\), where $N$ is the state dimension, and the parameterized functions \(S_{\Delta,d}:\mathbb{R}^D\rightarrow \mathbb{R}\), \(S_{\mathbf{B}}:\mathbb{R}^D\rightarrow \mathbb{R}^N\), and \(S_{\mathbf{C}}:\mathbb{R}^D\rightarrow \mathbb{R}^N\), which are implemented as linear transformations, are learned during training~\cite{gu2023mamba}. % To ensure stability, \(A_d\) is typically constrained to be diagonal with strictly negative eigenvalues.

% where \( \Delta_d (u) = \text{softplus}(S_{\Delta, d}(u))\), with $u \in \mathbb{R}^D$, is time-step scaling factor. The hidden matrices \( A_d \in \mathbb{R}^{N \times N} \), along with the parameterized functions \( S_{\Delta, d}(\cdot) \), \( S_{\mathbf{B}}(\cdot) \), and \( S_{\mathbf{C}}(\cdot) \), which are implemented as linear transformations, are all learned from the data during training.
% To ensure stability, \(A_d\) is typically constrained to be diagonal with strictly negative eigenvalues.

\subsubsection{Human Motion Representation.} Human motion data captures both temporal and spatial dynamics. A motion sequence is represented as $\mathbf{X} = (X_1, X_2, \ldots, X_L) \in \mathbb{R}^{L \times D}$, where $L$ is the sequence length and $D$ denotes the spatial dimension of each pose. Each pose $x_l$ is represented in the SMPL format~\citep{loper2023smpl} and contains information such as joint rotations, positions, or velocities.

\vspace{-1ex}
\subsection{Problem Definition}
\label{sec:problem}
%\textbf{Conditioned Temporal Mamba for Human Motion.} 
Given the $L$-length condition embedding \(\mathbf{m} = (m_1, m_2, \ldots, m_L) \in \mathbb{R}^{L\times E}\), our goal is to generate/estimate a human motion sequence \(\mathbf{X} = (X_1, X_2, \ldots, X_L) \in \mathbb{R}^{L\times D}\) that is temporally aligned with the given condition. We note that the condition input $\mathbf{m}$ and the human motion sequence $\mathbf{X}$ are assumed to \textit{have the same length} $L$. In practice, the condition \(\mathbf{m}\) represents temporal inputs such as music, egocentric video, or sequences of object geometry. % to generate the human motion. In this paper, we propose a new method that directly injects the conditioning information into the Mamba block, enabling more effective and fine-grained integration of multi-modal signals. 

Following previous works~\citep{tseng2023edge, tevet2023human}, we adopt a diffusion model~\citep{ho2020denoising} as a backbone framework due to its superior results in various tasks~\cite{tseng2023edge,siyao2022bailando,zhu2023human}. The diffusion model includes a forward process and a backward process. In the forward process, clean motion data \(\mathbf{X}_0\) is progressively added with Gaussian noise over $t$ time steps, resulting in noisy motion \( \mathbf{X}_t \). This process can be formulated as:
\begin{equation}\label{eq: forward}
 q(\mathbf{X}_t | \mathbf{X}_0) = \mathcal{N}( \sqrt{\alpha_t} \mathbf{X}_0, (1 - \alpha_t) \mathbf{I} )\Comma
\end{equation}
where \( \alpha_t = \prod_{s=1}^{t} (1 - \delta_s) \), and \( \delta_t \) controls the noise schedule. In the reverse process, a mamba-based neural network $f_\theta$ is learned to progressively remove noise from the noisy motion sequence $\mathbf{X}_t$ and recover the original clean motion $\mathbf{X}_0$, conditioned on an auxiliary modality $\mathbf{m}$. Following the approach of~\cite{ho2020denoising}, we adopt the training objective:
\begin{equation} 
\hat{\theta} = \arg\min_{\theta} \mathbb{E}_{t, \mathbf{X}_t} \left[ \|\mathbf{X}_0 - f_\theta (\mathbf{X}_t, t, \mathbf{m})\| \right]\FullStop
\end{equation}

\subsection{Temporally Conditional Mamba for Human Motion}
\label{sec:tcm}
\subsubsection{Overview.} We introduce a new mamba-based diffusion model for learning human motion. Our design is a mamba-based architecture where the condition embedding is directly fused within the mamba block, particularly in a temporally-aware manner. Fig.~\ref{fig: method} shows an overview of the proposed architecture. The model takes as input the noised motion \(\mathbf{x}\), the condition embedding \(\mathbf{m}\), and the timestep embedding \(\mathbf{t}\), and predicts the denoised motion \(\hat{\mathbf{x}}\). The architecture consists of three main components: (1) \textit{Temporally Conditional Mamba}, which injects the condition modality embedding at the token level to align the motion with the conditioning signal temporally; (2) \textit{Spatial Mamba}, which captures spatial dependencies of human motion; and (3) \textit{Adaptive Layer Norm}, which globally incorporates both the condition embedding and timestep embedding uniformly across all tokens. In the following, we provide a detailed description of each model component. 

\subsubsection{Temporally Conditional Mamba.} In control theory, Linear Parameter-varying State Space Models~\citep{mohammadpour2012control, briat2014linear} adapt system matrices, such as $\mathbf{B}$ and $\mathbf{C}$, dynamically based on external signals, allowing the system to adjust its internal dynamics and output behavior flexibly. Motivated by this principle, we propose Temporally Conditional Mamba (TCM), where the condition signal is used to modulate the \( \mathbf{B} \) and \( \mathbf{C} \) matrices over time; this mechanism intuitively enables the model to dynamically tailor its evolution at each frame, leading to outputs that more tightly aligned with the temporal patterns present in the conditioning information. The detailed procedure of TCM is presented in Algorithm~\ref{algorithm}. It transforms a sequence of temporal tokens embedding $\mathbf{x}\in \mathbb{R}^{B\times L\times D}$ with the input condition embedding $\mathbf{m}\in \mathbb{R}^{B\times L\times E}$ to output fused motion-condition embedding $\mathbf{y}\in\mathbb{R}^{B\times L\times D}$, where, $B$ denotes the batch size, $L$ is the temporal length, and $D,E$ are the embedding dimensions. Specifically, the selection matrices $\mathbf{B}$ and $\mathbf{C}$ are modified to depend on both motion embedding $\mathbf{x}$ and condition embedding $\mathbf{m}$. The update rule in Equation~\ref{eq: origin_mamba} is redefined as:
% We adopt most similar structure of the original Mamba block~\citep{gu2023mamba}, except the 'selection' matrices $\mathbf{B}$ and $\mathbf{C}$, which we modify to depend on both motion embedding $\mathbf{x}$ and condition embedding $\mathbf{m}$ as follows:
\vspace{-1ex}
\begin{equation} \label{eq: ca_mamba}
    \begin{aligned} 
        h_{dl} &= \bar{\mathbf{A}}_{dl} \cdot h_{d,l-1} + 
        (\Delta_{dl}\cdot\widetilde{\mathbf{B}}_{l}(x_l,m_l)) \cdot x_{dl}
        % , \quad h_{d0} = [m_1,0^{(L-1)}]
        \Comma \\ 
        y_{dl} &= \widetilde{\mathbf{C}}_l(x_l,m_l) \cdot h_{dl}\Comma
    \end{aligned}
\end{equation}
for $l=1,2,\ldots,L, \text{and}~d=1,2,\ldots,D$. Here, \(\widetilde{\mathbf{B}}_l\) and \(\widetilde{\mathbf{C}}_l\) are condition-aware selection matrices. They are obtained via an element-wise affine modulation of the original matrices:
\begin{equation}
    \begin{aligned}
        % &\gamma_{\mathbf{B}}, \beta_{\mathbf{B}}, \gamma_{\mathbf{C}},\beta_{\mathbf{C}} = S_m(\mathbf{m})\Comma \\
        &\mathbf{\widetilde{B}}_l = \gamma_{\mathbf{B}}(m_l)\odot \mathbf{B}_l + \beta_{\mathbf{B}}(m_l)\Comma \\
        &\mathbf{\widetilde{C}}_l = \gamma_{\mathbf{C}}(m_l)\odot \mathbf{C}_l + \beta_{\mathbf{C}}(m_l)\Comma
    \end{aligned}
\end{equation}
where \(\gamma_{\mathbf{B}}, \beta_{\mathbf{B}}, \gamma_{\mathbf{C}}, \beta_{\mathbf{C}} \in \mathbb{R}^N\), where $N$ is state dimension, are predicted by a learnable mapping \(S_m: \mathbb{R}^E \to \mathbb{R}^{N}\) applied to \(m_l\).  
% For computational efficiency, \(S_m(\cdot)\) is instantiated as a single linear layer followed by a non-linear activation (Figure~\ref{fig: method}b).
% where, \(\gamma_{\mathbf{B}}, \beta_{\mathbf{B}}, \gamma_{\mathbf{C}}, \beta_{\mathbf{C}}\) are modulation parameters produced by a learnable mapping \(S_m(\cdot)\) from the condition embedding \(\mathbf{m}\).  
For simplicity and computational efficiency, we implement \(S_m(\cdot)\) as a single linear layer with a non-linear activation, as illustrated in Fig.~\ref{fig: method}b.  
%This mechanism can be viewed as a form of Feature-wise Linear Modulation (FiLM)~\citep{perez2018film}, where the condition embedding generates affine transformation parameters that modulate the internal dynamics of the model in a token-wise manner. 
By making the selection matrices \(\mathbf{B}\) and \(\mathbf{C}\) dynamically dependent on \(\mathbf{m}\), the model is endowed with fine-grained control over how the input sequence \(\mathbf{x}\) and condition \(\mathbf{m}\) interact to produce the output \(\mathbf{y}\).  
This mechanism allows the model to adapt its recurrent dynamics based on the external condition.

\begin{algorithm}[h]
\LinesNumbered
\caption{Temporally Conditional Mamba} \label{algorithm}
\KwIn{Motion embedding $\mathbf{x}: (B,L,D)$, \\ Condition embedding $\mathbf{m}: (B, L, E)$}
\KwOut{Transformed motion $\mathbf{y}: (B, L, D)$}

\textit{/* Input projection */} \;
$\mathbf{u} \gets S_{\mathbf{u}}(\mathbf{x}) \quad : (B, L, D)$\;
$\mathbf{z} \gets S_{\mathbf{z}}(\mathbf{x}) \quad : (B, L, D)$\;
$\mathbf{u} \gets \text{SiLU}(\text{Conv1d}(\mathbf{u})) \quad :(B, L, D)$\;

$\mathbf{A} \gets \text{Parameter} \quad : (D, N)$\;

\textit{/* Get $\mathbf{B}$ and $\mathbf{C}$ */} \;
$\mathbf{B} \gets S_{\mathbf{B}}(\mathbf{u}) \quad : (B, L, N)$\;
$\mathbf{C} \gets S_{\mathbf{C}}(\mathbf{u}) \quad : (B, L, N)$\;

\textit{/* Get condition-aware $\widetilde{\mathbf{B}}$ and $\widetilde{\mathbf{C}}$ */} \;
$(\gamma_{\mathbf{B}}, \beta_{\mathbf{B}}, \gamma_{\mathbf{C}}, \beta_{\mathbf{C}}) \gets S_m(\mathbf{m}) \quad : (B, L, N)$\;
$\widetilde{\mathbf{B}} \gets \gamma_{\mathbf{B}} \odot \mathbf{B} + \beta_{\mathbf{B}} \quad : (B, L, N)$\;
$\widetilde{\mathbf{C}} \gets \gamma_{\mathbf{C}} \odot \mathbf{C} + \beta_{\mathbf{C}} \quad : (B, L, N)$\;

\textit{/* Get discrete $\bar{\mathbf{B}}$ and $\bar{\mathbf{A}}$ */} \;
$\Delta \gets \text{softplus}(S_\Delta(\mathbf{u})) \quad : (B, L, D)$\;
$\bar{\mathbf{B}} \gets \Delta \cdot \widetilde{\mathbf{B}} \quad : (B, L, D, N)$\;
$\bar{\mathbf{A}} \gets \exp(\Delta \cdot \mathbf{A}) \quad : (D, N)$\;

\textit{/* SSM forward */} \;
$\mathbf{y} \gets \text{SSM}(\bar{\mathbf{A}}, \bar{\mathbf{B}}, \widetilde{\mathbf{C}})(\mathbf{u}) \quad : (B, L, D)$\;

\textit{/* Gating and output projection */} \;
$\mathbf{y} \gets S_\mathbf{y}(\mathbf{y} \odot \text{SiLU}(\mathbf{z})) \quad : (B, L, D)$\;

\Return $\mathbf{y}$
\end{algorithm}

\subsubsection{Spatial Mamba.} Apart from temporal dynamics, learning spatial dependencies among joints is critical for human motion generation, as spatial structures encode essential information complementary to temporal dynamics. A Spatial Mamba block is used to model spatial dependencies. The detailed implementation of Spatial Mamba is provided in Fig.~\ref{fig: method}c. First, the motion representation is rearranged from the temporal domain \((B, L, D)\) to the spatial domain \((B, D, L)\). A standard Mamba block~\citep{gu2023mamba} is then applied to model interactions across joints. Finally, the spatially transformed sequence is rearranged back to the original temporal format. This design enables the model to reason about both spatial-temporal structures in human motion effectively.

\subsubsection{Adaptive Layer Norm.} 
% To further enhance \textbf{conditional injection}, \textit{shoud avoid using this, conflict with the contribution about TCM. Maybe rewrite to stress about ``normalization" (instead of ``condition"?)}
% Following previous works that use Adaptive Layer Norm (AdaLN) in diffusion models, we apply a adaptive normalization layer (AdaLN)~\citep{karras2019style, peebles2023scalable} that replace the standard layer norm. Specifically, dimension-wise scale and shift parameters \(\gamma\) and \(\beta\) are computed from the sum of the timestep embedding and condition embedding via an MLP. The motion embedding $\mathbf{x}$ is then transformed as:
% \begin{equation}
% \mathbf{x}' = \gamma(\mathbf{m},\mathbf{t}) \odot \text{Norm}(\mathbf{x}) + \beta(\mathbf{m},\mathbf{t})
% \end{equation}
% We apply this block before This adaptive modulation is applied uniformly across all tokens in the sequence. Together with the token-wise conditioning in Temporally Condition Mamba, this design enables both fine-grained and global alignment between the generated motion and the conditioning inputs.
Following prior work on adaptive normalization in diffusion models~\cite{dhariwal2021diffusion, peebles2023scalable}, we incorporate an Adaptive Layer Normalization (AdaLN) layer~\citep{karras2019style, peebles2023scalable} to replace the standard LayerNorm. AdaLN effectively integrates timestep and condition information by modulating the feature distribution based on their combined embedding. Specifically, dimension-wise scale and shift parameters $\lambda_i$ and $\rho_i$ are generated from the sum of the timestep embedding $\mathbf{t}$ and condition embedding $\mathbf{m}$ through an MLP:
\begin{equation}
    \lambda_i, \rho_i = \text{MLP}(\text{Sum}(\mathbf{t},\mathbf{m}))\Comma
\end{equation}
where $i \in \{S, T\}$, with $S$ referring to the spatial and $T$ to the temporal normalization. The motion embedding $\mathbf{x}$ is then modulated as:
\begin{equation}
    \mathbf{x}' = \lambda_i \odot \text{Norm}(\mathbf{x}) + \rho_i\FullStop
\end{equation}
We apply AdaLN before both the Spatial Mamba and Temporally Conditional Mamba blocks. This mechanism operates uniformly across all tokens in the sequence.
Combined with the token-wise conditioning inside Temporally Conditional Mamba, this design improves the alignment between the generated motion and the conditioning inputs across the entire sequence.
% \textbf{need some rewrite. ``global condition" may not good}

\section{Experiment}
\label{sec:exp}
In this section, we first conduct a standard task setup to compare our method with the Vanilla Mamba and Cross-Attention approaches. We then demonstrate the generalization of our proposed method in four popular human motion learning tasks, all of which involve temporally aligned conditional inputs. The four tasks include: (\textit{i}) human motion synthesis from music, (\textit{ii}) human motion estimation from egocentric video, (\textit{iii}) human motion estimation from egocentric video and music, and (\textit{iv}) human motion synthesis from object trajectories. These tasks cover both human motion \textit{synthesis} and human motion \textit{estimation} settings.

\subsection{Comparison with Vanilla Mamba and Cross-Attention}
\label{subsec_basic_comp}

% \textbf{visualization to compare valila mamba and cross-attenion did not align with temporal condition - beat music - need to stress/swing hand}

To make a fair comparison between our method and traditional approaches, we first set up a simple experiment that focuses only on techniques used in fusing the temporal condition to generate human motion. We select human dance synthesis from music as the primary task for this experiment, as it is a widely studied and representative benchmark in motion synthesis~\cite{zhu2023human}.

\subsubsection{Implementation.} We implement a diffusion-based framework~\cite{ho2020denoising} in which music serves as the conditioning signal for all methods: our TCM, Vanilla Mamba, and Cross-Attention. Following~\cite{tseng2023edge}, we use Jukebox~\citep{dhariwal2020jukebox}, a GPT-style pretrained generative model, to extract a 4800-dim feature that effectively captures the characteristics of the input music. In Vanilla Mamba, we replace our TCM block with standard Mamba blocks~\cite{gu2023mamba}; conditioning is integrated solely through adaptive layer normalization in this setting. In the Cross-Attention setup, we build upon the Vanilla Mamba setup by inserting a cross-attention module into each layer for condition injection.

\subsubsection{Dataset and Metrics.} We utilize the AIST++ dataset~\citep{li2021ai}, which contains 1,408 high-quality dance motion sequences paired with music spanning a wide range of genres. We adopt the original training and testing splits provided by the dataset. To assess the quality of the generated dance motions, we use \(\text{FID}_k\) and \(\text{FID}_g\) to measure fidelity, and \(\text{Div}_k\) and \(\text{Div}_g\) to assess diversity of the generated motion, where \(k\) and \(g\) refer to the kinematic and geometric feature of the generated motion, respectively. We employ the Beat Alignment Score (BAS)~\cite{siyao2022bailando} to quantify how well the generated motion aligns with the input music.

\begin{table}[]
    \centering
    \setlength{\tabcolsep}{0.5ex}
    \renewcommand\arraystretch{1.2}
    \caption{Comparative results of dance synthesis from music task. We compare our proposed TCM with Cross-Attention and Vanilla Mamba. Bold indicates best, and underline indicates second best.}
    \vspace{-2ex}
    \label{tab: compare_to_ca}
    \resizebox{1\linewidth}{!}{
    \begin{tabular}{@{}lccccccc@{}}
        \toprule
        Method & Params (M) & Infer (s) &$\text{FID}_k\downarrow$ & $\text{FID}_g\downarrow$ & $\text{Div}_k\uparrow$ & $\text{Div}_g\uparrow$ & BAS~$\uparrow$ \\ \midrule
        Ground Truth & - & - & 17.10 & 10.60 & 8.19 & 7.45 & 0.2374 \\ \midrule
        Vanilla Mamba & 26.58 & 1.04 & 25.61 & 16.35 & 7.52 & 6.05 & \underline{0.2434} \\
        Cross-Attention & 41.28 & 1.42 & \underline{23.43} & \underline{12.86} & \underline{7.87} & \underline{6.48} & 0.2411 \\
        \textbf{TCM} (ours) & 26.84 & 1.11 & \textbf{20.66} & \textbf{9.75} & \textbf{8.98} & \textbf{7.24} & \textbf{0.2761} \\ \hline
    \end{tabular}}
\end{table}

\begin{table}[]
    \centering
    \setlength{\tabcolsep}{2ex}
    \caption{Ablation experiment on network components. We assess the performance of the proposed method under different settings.}
    \vspace{-2ex}
    \label{tab: ablation}
    \resizebox{\linewidth}{!}{
    \begin{tabular}{@{}lccccc@{}}
        \toprule
         & $\text{FID}_k\downarrow$ & $\text{FID}_g\downarrow$ & $\text{Div}_k\uparrow$ & $\text{Div}_g\uparrow$ & BAS~$\uparrow$ \\ \midrule
        w.o. TCM block & 25.61 & 16.35 & 7.52 & 6.05 & 0.2434 \\
        w.o. AdaLN block & 21.89 & 11.76 & 8.11 & 6.33 & 0.2615 \\
        w.o. $\gamma_{\mathbf{B},\mathbf{C}}$ & 22.37 & 11.48 & 8.02 & 6.36 & 0.2582 \\
        w.o. $\beta_{\mathbf{B},\mathbf{C}}$ & 21.52 & 10.63 & 8.34 & 6.58 & 0.2624 \\
        Best architecture & \textbf{20.66} & \textbf{9.75} & \textbf{8.98} & \textbf{7.24} & \textbf{0.2761} \\ \bottomrule
    \end{tabular}}
\end{table}

\begin{table}[]
    \centering
    \caption{Ablation experiments on sequence length of motion. Length analysis. We evaluate the FID score of our TCM and Cross-Attention at different sequence lengths.}
    \begin{tabular}{lccc}
    \toprule
    Motion length (s) & 5 & 15 & 25 \\ \midrule
    Cross-Attention & 25.61 & 27.04 & 31.36  \\
    \textbf{TCM} (ours) & 20.66 & 21.09 & 23.82 \\ \bottomrule
    \end{tabular}
    \label{tab:length_anal}
\end{table}

\subsubsection{Main Results.} 
\begin{figure}
    \centering
    \includegraphics[width=\linewidth]{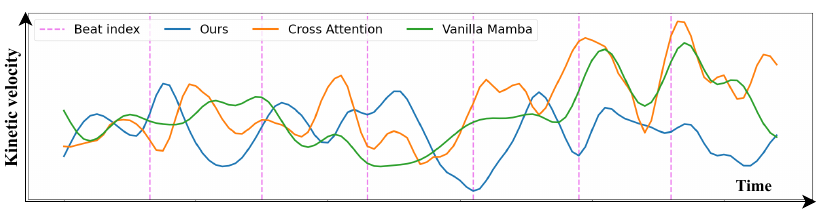}
    \caption{Motion and music beat alignment. We plot the mean joint velocity over time for different methods. Kinematic beats are identified as local minima in the velocity curves. Our method produces motion with kinematic beats that align more closely with the music beats.}
    \label{fig: bas_vis}
\end{figure}

Table~\ref{tab: compare_to_ca} shows that our TCM outperforms both Vanilla Mamba and Cross-Attention. We also report model parameters and inference time (evaluated using DDIM with 50 sampling steps). Our method achieves better performance while maintaining a parameter count and inference speed comparable to that of Vanilla Mamba. Although the Cross-Attention model demonstrates slightly higher motion realism than Vanilla Mamba, it exhibits weaker synchronization with the music inputs. In Fig.~\ref{fig: bas_vis}, we show the mean velocity of generated motion over time. We can see that our generated dance has motion beats,  identified as local minima in kinetic velocity, that closely match the music beats, while both Cross-Attention and Vanilla Mamba fail. Table~\ref{tab: compare_to_ca} and Fig.~\ref{fig: bas_vis} suggest that autoregressively injecting the condition into the motion sequence enables more precise and temporally consistent alignment with the conditioning signal, particularly for temporal data where time-dependent correlations are critical.

\subsection{Ablation analysis}
\subsubsection{The effect of each component.} 
To understand the contributions of each component to the overall results, we conduct an analysis in Table~\ref{tab: ablation}. The results indicate that with our TCM block, the BAS score improves from 0.2423 to 0.2761, and the same trend is observed for other metrics. Additionally, integrating AdaLN layers yields slight performance gains. These results confirm the TCM block's key role in aligning generated motion with music.  Table~\ref{tab: ablation} also shows the impact of scale and shift parameters in the TCM block. We find that the performance drop when not using $\gamma$ is higher than when not using $\beta$. This result suggests that the scale parameter $\gamma$ plays a more critical role than the shift parameter $\beta$.

\subsubsection{Could TCM distinguish input conditions?} 
\begin{figure}
\centering
\includegraphics[width=\linewidth]{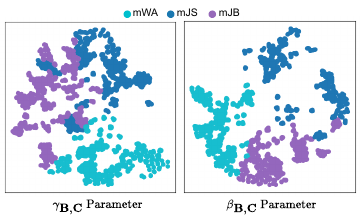}
    \caption{t-SNE visualization of $\gamma_{\mathbf{B},\mathbf{C}}$ and $\beta_{\mathbf{B},\mathbf{C}}$ in TCM blocks. Clustering in the t-SNE plot reveals that the learned $\gamma_{\mathbf{B},\mathbf{C}}$ and $\beta_{\mathbf{B},\mathbf{C}}$ parameters encode discriminative information that reflects distinct music genres.}
    \label{fig: tnse}
\end{figure}

% (explain how we visualize/get that image, explain what inside those images, discuss the message/outcomes of that image bring)
To gain insights into how our TCM processes conditional information, we investigate learned  $\gamma$ and $\beta$ parameters in the TCM block. We extract $\gamma_{\mathbf{B}},\beta_{\mathbf{B}},\gamma_{\mathbf{C}},\beta_{\mathbf{C}}$ from 5000 music samples across three distinct music genres (i.e., mWA, mJS, mJB). We visualize the t-SNE plot of these parameters in Fig.~\ref{fig: tnse}. The results reveal clear clustering patterns based on music genre, indicating that the TCM adapts its internal dynamics to different conditioning signals.    

\subsubsection{Sequence Length Analysis.} To analyze the effect of sequence length, we evaluate the FID score at 5, 15, and 25 seconds. We compare our TCM with the Cross-Attention baseline. As shown in Table~\ref{tab:length_anal}, the FID score of motions generated by our method increases more slowly as the sequence length grows, indicating superior capability for long-range motion generation.

\vspace{-1ex}
\subsection{Comparison with State-of-the-art Methods across Different Tasks}
\subsubsection{Human Dance Synthesis from Music} 
Human dance synthesis from music is a prominent task with wide-ranging applications in animation, film production~\citep{chan2019everybody}, and the metaverse~\citep{lam2022human}. The goal is to generate realistic dance movements that are temporally synchronized with a given piece of music. As in Section~\ref{subsec_basic_comp}, our model, which incorporates condition-aware temporal modeling, is particularly well-suited for this task. 

\textbf{Results.} We compare our proposed method, TCM, with several state-of-the-art methods on the AIST++ dataset. These baselines adopt a variety of condition fusion mechanisms, ranging from simple concatenation to cross-attention, as detailed in Table~\ref{tab: music2motion}. The results show that our model outperforms previous approaches, demonstrating better motion quality and diversity. Notably, our TCM achieves a BAS metric of 0.2761, indicating a stronger temporal alignment between the generated dance and music.

\subsubsection{Human Motion Estimation from Egocentric Video}
Estimating human motion from egocentric video captured by a single head-mounted camera requires understanding of physically plausible full-body motion, with a strong emphasis on aligning head movements to the video on a frame-by-frame basis. We integrate our TCM into the framework in ~\cite{li2023ego}, replacing its transformer backbone with our TCM. For evaluation, we compare our approach with four baselines: PoseReg~\citep{yuan2019ego}, KinPoly-OF~\citep{luo2021dynamics}, AvatarPoser~\citep{jiang2022avatarposer}, and EgoEgo~\citep{li2023ego}.

\begin{figure*}[ht]
    \centering \includegraphics[width=\linewidth]{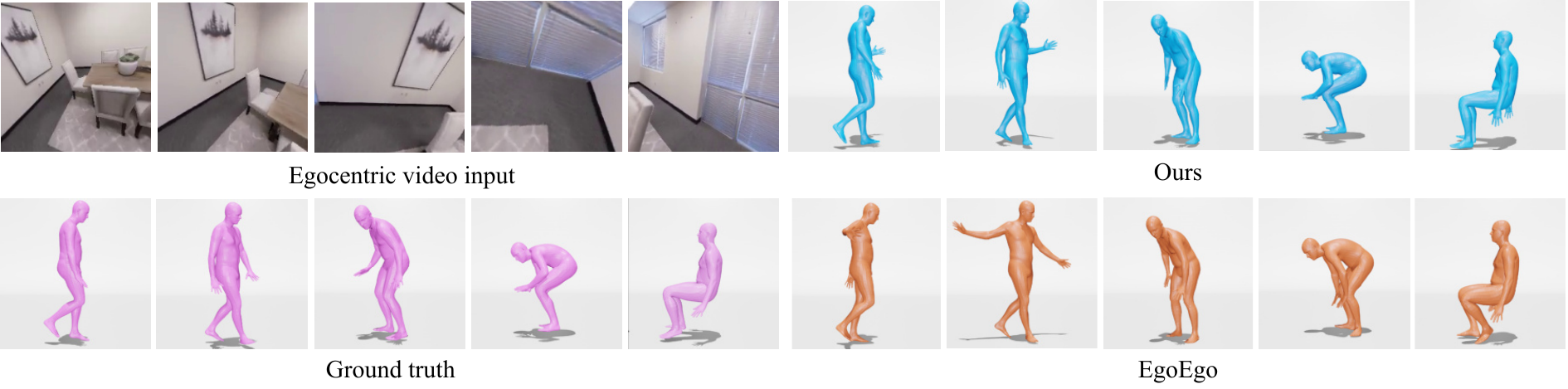}
    \vspace{-2ex}\caption{Qualitative comparison of human motion estimation from the egocentric task. Our method produces more coherent and accurate motion compared to EgoEgo. For additional visualizations, please refer to our demo video.}
    \label{fig: ego2motion}
    \vspace{-1ex}
\end{figure*}

\begin{figure*}[t]
    \centering
    \subfloat[Ground truth]{\includegraphics[width=0.33\linewidth]{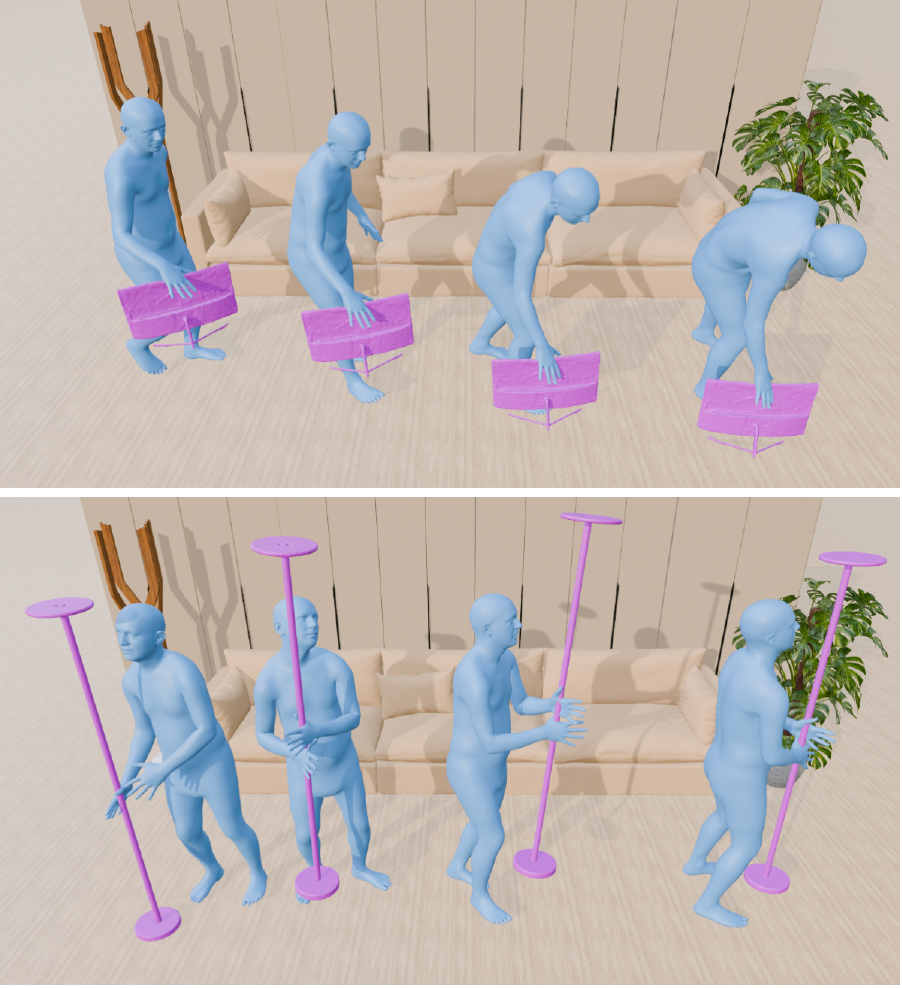}}
    \hfill
    \subfloat[OMOMO]{\includegraphics[width=0.33\linewidth]{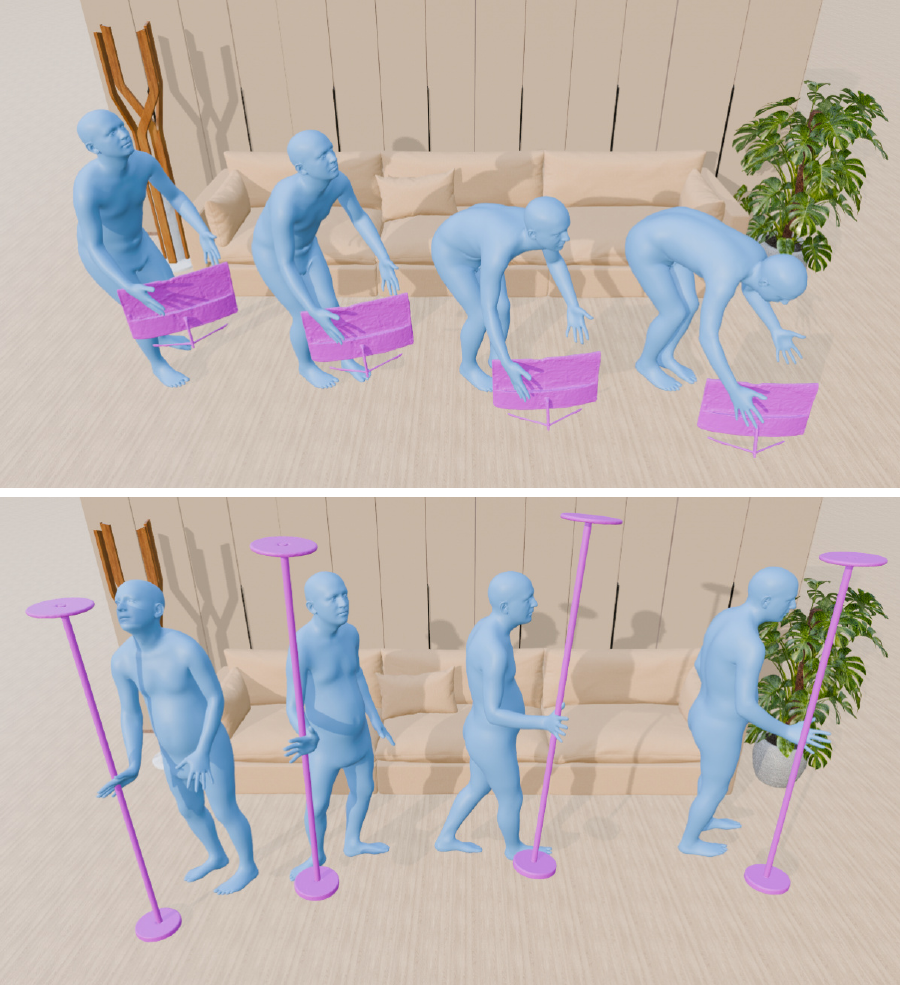}}
    \hfill
    \subfloat[Ours]{\includegraphics[width=0.33\linewidth]{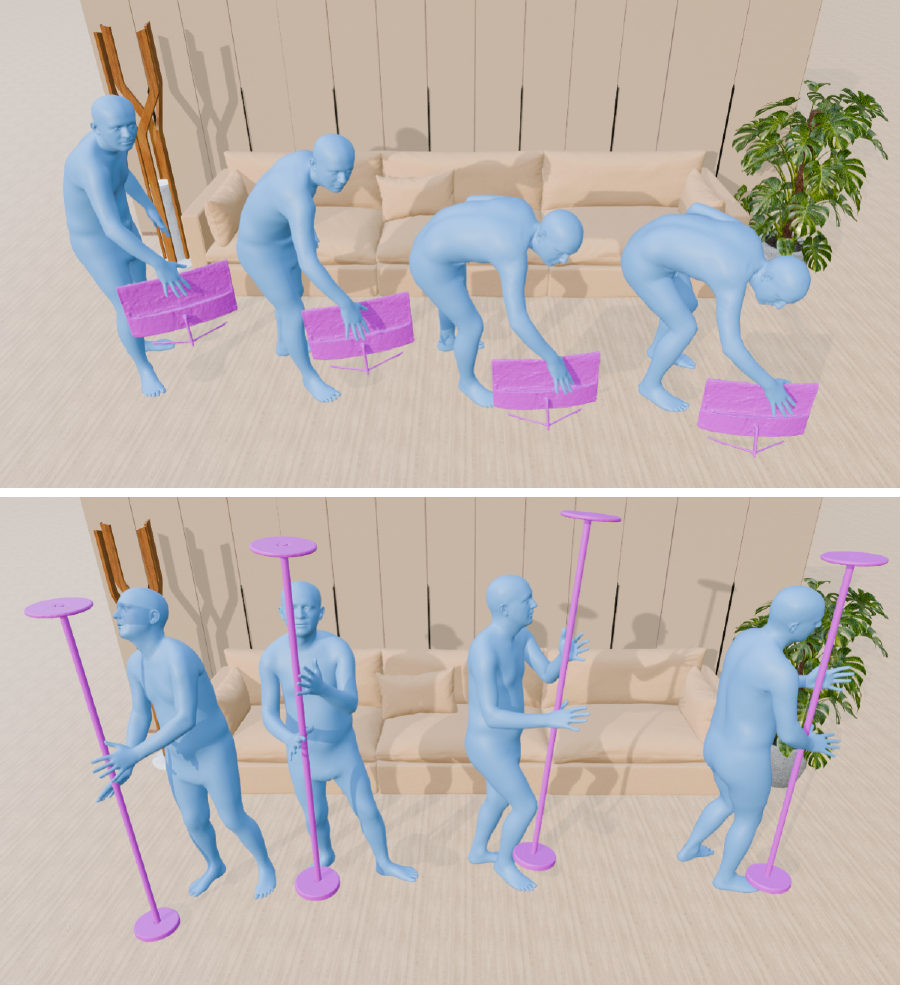}}
    \vspace{-2ex}
    \caption{Qualitative comparison on human motion generation from object trajectories. The top row shows an example where OMOMO produces motion that resembles carrying a box rather than a monitor. In the bottom example, the contact generated by our method appears more plausible compared to OMOMO.}
    \label{fig:object2motion}
    \vspace{-1ex}
\end{figure*}

\begin{table}[]
    \centering
    \caption{Comparison with State-of-the-art methods on music-to-dance task. Bold indicates the best results, and underlined indicates the second-best results.}
    \label{tab: music2motion}
    \resizebox{\linewidth}{!}{
    \begin{tabular}{@{}lccccc@{}} \toprule
    \multirow{2}{*}{Method} & \multicolumn{2}{c}{Motion Quality} & \multicolumn{2}{c}{Motion Diversity} & \multirow{2}{*}{BAS$\uparrow$} \\ \cmidrule(r){2-3} \cmidrule(r){4-5}
     & $\text{FID}_k\downarrow$ & $\text{FID}_g\downarrow$ & $\text{Div}_k\uparrow$ & $\text{Div}_g\uparrow$ &  \\  \midrule
    Ground Truth & 17.10 & 10.60 & 8.19 & 7.45 & 0.2374 \\ \midrule 
    TSMT~\citep{li2020learning} & 86.43 & 43.46 & 6.85 & 3.32 & 0.1607 \\
    DanceNet~\citep{zhuang2022music2dance} & 69.18 & 25.49 & 2.86 & 2.85 & 0.1430 \\
    DanceRev~\citep{huang2020dance} & 73.42 & 25.92 & 3.52 & 4.87 & 0.1950 \\
    FACT~\citep{li2021ai} & 35.35 & 22.11 & 5.94 & 6.18 & 0.2209 \\
    Bailando~\citep{siyao2022bailando} & \underline{28.16} & \textbf{9.62} & 7.83 & 6.34 & 0.2332 \\
    EDGE~\citep{tseng2023edge} & 42.16 & 22.12 & 3.96 & 4.61 & 0.2334 \\
    LDA~\cite{alexanderson2023listen} & 25.67 & 11.25 & 8.02 & 6,58 & \underline{0.2431} \\
    BADM~\citep{zhang2024bidirectional} & - & - & \underline{8.29} & \underline{6.76} & 0.2366 \\
    Lodge~\citep{li2024lodge} & 37.09 & 18.79 & 5.58 & 4.85 & 0.2423 \\
    \textbf{TCM} (ours) & \textbf{20.66} & \underline{9.75} & \textbf{8.98} & \textbf{7.24} & \textbf{0.2761} \\ \bottomrule
    \end{tabular}}
\end{table}

\begin{table}[]
    \centering
      \caption{Comparison with State-of-the-art methods on ego-to-motion task. Bold indicates the best, and underline indicates the second-best results.}
    \label{tab: ego2motion_re}
    \resizebox{\linewidth}{!}{
\begin{tabular}{lccccc} \toprule
        Method & $\mathbf{O}_{\text{head}}$$\downarrow$ & $\mathbf{T}_{\text{head}}$$\downarrow$ & MPJPE$\downarrow$ & Accel$\downarrow$ & FS$\downarrow$ \\ \midrule
        PoseReg~\citep{yuan2019ego} & 0.77 & 354.7 & 147.7 & 127.6 & 87.1 \\
        Kinpoly-OF~\citep{luo2021dynamics} & 0.62 & 323.4 & 141.6 & 7.3 & 4.2 \\
        AvatarPoser~\citep{jiang2022avatarposer} & 0.43 & \underline{140.2} & 126.5 & 7.1 & 3.8 \\
        EgoEgo~\citep{li2023ego} & \underline{0.20} & 148.0 & \underline{121.1} & 6.2 & \underline{2.7} \\
        \textbf{TCM} (ours) & \textbf{0.15} & \textbf{112.6} & \textbf{116.3} & \textbf{6.2} & \textbf{2.4} \\ \bottomrule
        \end{tabular}}
\end{table}
% \begin{table}[]
%     \centering
%     \setlength{\tabcolsep}{0.6ex}
%     \caption{Performance comparison between proposed and other methods on human dance estimation from egocentric and music. Bold indicates best results and underline indicates second best results.}
%     \label{tab: egomusic2motion}
%     \resizebox{\linewidth}{!}{
%     \begin{tabular}{lcccccc}
%     \toprule
%     Baselines & $\textbf{O}_{head}$$\downarrow$ & $\textbf{T}_{head}$$\downarrow$ & MPJPE$\downarrow$ & Accel$\downarrow$ & FS$\downarrow$ & BVAS$\uparrow$ \\ \midrule
%     Pose-Reg~\cite{yuan2019ego} & 1.21 & 642.47 & 377.56 & 30.36 & 56.18 & 0.165 \\
%     Kinpoly~\cite{luo2021dynamics} & 0.78 & 354.19 & 251.27 & 17.84 & 25.31 & 0.187 \\
%     EgoEgo~\cite{li2023ego} & \underline{0.67} & \underline{347.23} & 234.58 & 16.76 & 20.15 & 0.203 \\
%     FACT~\cite{li2021ai} & 1.37 & 685.81 & 244.89 & 17.54 & 18.53 & 0.195 \\
%     Bailando~\cite{siyao2022bailando} & 1.44 & 688.54 & 231.77 & 14.23 & 18.67 & 0.211 \\
%     EDGE~\cite{tseng2023edge} & 1.27 & 644.62 & \underline{213.37} & \underline{13.78} & \underline{14.33} & \underline{0.221} \\
%     \textbf{TCM} (ours) & \textbf{0.50} & \textbf{335.41} & \textbf{130.22} & \textbf{11.05} & \textbf{12.34} & \textbf{0.265} \\ \bottomrule
%     \end{tabular}}
%     \vspace{-2.5ex}
% \end{table}

\begin{table}[]
    \centering
    \setlength{\tabcolsep}{0.6ex}
    \caption{Performance comparison between proposed and other methods on human dance estimation from egocentric and music. Bold indicates the best results, and underline indicates the second-best results.}
    \label{tab: egomusic2motion}
    \resizebox{\linewidth}{!}{
    \begin{tabular}{lccccc}
    \toprule
    Baselines & $\textbf{O}_{head}$$\downarrow$ & $\textbf{T}_{head}$$\downarrow$ & MPJPE$\downarrow$ & Accel$\downarrow$ & FS$\downarrow$ \\ \midrule
    Pose-Reg~\cite{yuan2019ego} & 1.21 & 642.47 & 377.56 & 30.36 & 56.18 \\
    Kinpoly~\cite{luo2021dynamics} & 0.78 & 354.19 & 251.27 & 17.84 & 25.31 \\
    EgoEgo~\cite{li2023ego} & \underline{0.67} & 347.23 & 234.58 & 16.76 & 20.15 \\
    FACT~\cite{li2021ai} & 1.37 & 685.81 & 244.89 & 17.54 & 18.53  \\
    Bailando~\cite{siyao2022bailando} & 1.44 & 688.54 & 231.77 & 14.23 & 18.67  \\
    EDGE~\cite{tseng2023edge} & 1.27 & 644.62 & \underline{213.37} & \underline{13.78} & \underline{14.33} \\
    EMM~\cite{nguyen2025egomusic} & 0.61 & \textbf{322.19} & 191.55 & 12.76 & 13.18 \\
    \textbf{TCM} (ours) & \textbf{0.50} & \underline{335.41} & \textbf{130.22} & \textbf{11.05} & \textbf{12.34} \\ \bottomrule
    \end{tabular}}
    \vspace{-2.5ex}
\end{table}

\textbf{Dataset and metrics.} We use ARES dataset~\citep{li2023ego}, which comprises approximately 2,354 indoor motion sequences paired with corresponding egocentric videos. To evaluate the results, we adopt the evaluation metrics used in EgoEgo~\citep{li2023ego}, including: mean per joint position error (MPJPE) for overall positional accuracy; head pose error, measured by orientation (\(\textbf{O}_{\text{head}}\)) and translation (\(\textbf{T}_{\text{head}}\)), for assessing head alignment; acceleration error (Accel) to evaluate motion smoothness; and foot skating (FS) to measure artifacts in foot-ground contact consistency.

% \begin{table*}[]
%     \caption{Comparative results on human motion generation from object movement task. Bold indicates best, and underline indicates second best.}
%     \label{tab: object2motion}
%     \vspace{-1ex}
%     \resizebox{\linewidth}{!}{
%     \begin{tabular}{lccccccccc}
%     \toprule
%     Method & \multicolumn{1}{l}{Hand JPE $\downarrow$} & MPJPE $\downarrow$ & MPVPE $\downarrow$ & $\mathbf{T}_{\text{root}} \downarrow$ & $\mathbf{O}_{\text{root}} \downarrow$ & Collision $\downarrow$ & $\mathbf{C}_{\text{pred}} \uparrow$ & $\mathbf{C}_{\text{rec}} \uparrow$ & F1 Score $\uparrow$ \\ \midrule
%     GOAL~\citep{taheri2022goal} & 49.90 & 15.64 & 21.82 & 34.35 & 0.76 & \textbf{0.12} & \textbf{0.83} & 0.23 & 0.32 \\
%     OMOMO~\citep{li2023object} & \underline{24.01} & \underline{12.42} & \underline{16.67} & \underline{18.44} & \underline{0.50} & 0.22 & \underline{0.82} & \textbf{0.70} & \textbf{0.72} \\
%     \textbf{TCM} (ours) & \textbf{23.40} & \textbf{11.34} & \textbf{15.18} & \textbf{17.46} & \textbf{0.45} & \underline{0.22} & 0.81 & \underline{0.67} & \underline{0.70} \\ \bottomrule
%     \end{tabular}
%     }
%     \vspace{-1ex}
% \end{table*}

\textbf{Results.} Table~\ref{tab: ego2motion_re} compares our method and several baselines on the ARES dataset. Our approach consistently outperforms all baselines by a large margin across multiple metrics. 
% Furthermore, Fig.~\ref{fig: dtw_vis} visualizes the alignment between the predicted head trajectory and the ground-truth head trajectory using Dynamic Time Warping (DTW)~\cite{gold2018dynamic}. The results indicate that our method achieves higher trajectory similarity than EgoEgo, demonstrating its enhanced capacity to model realistic egocentric head motion. 
We present qualitative comparisons with the second-best method, EgoEgo, in Fig.~\ref{fig: ego2motion}. Our method produces more accurate and realistic motion, demonstrating better alignment with egocentric cues.

% \begin{table}[]
%     \caption{Comparative results on human motion generation from object movement task. Bold indicates best and underline indicates second best.}
%     \label{tab: object2motion}
%     \setlength{\tabcolsep}{0.5ex}
%     \resizebox{\linewidth}{!}{
%     \begin{tabular}{rcccc}
%     \toprule
%     Metrics & GOAL~\citep{taheri2022goal} & OMOMO~\citep{li2023object} & CHOIS~\citep{li2024controllable} & \textbf{TCM} (ours) \\ \midrule
%     Hand JPE $\downarrow$ & 49.90 & \underline24.01 & - & \textbf{23.40} \\
%     MPJPE $\downarrow$ & 15.64 & \underline12.42 & 15.82 & \textbf{11.34} \\
%     MPVPE $\downarrow$ & 21.82 & \underline16.67 & - & \textbf{15.18} \\
%     $\mathbf{T}_{\text{root}} \downarrow$ & 34.35 & 18.44 & 24.75 & \textbf{17.46} \\
%     $\mathbf{O}_{\text{root}} \downarrow$ & 0.76 & \underline0.50 & - & \textbf{0.45} \\
%     $\mathbf{C}_{\text{prec}} \uparrow$ & \textbf{0.83} & 0.82 & 0.77 & \underline0.82 \\
%     $\mathbf{C}_{\text{rec}} \uparrow$ & 0.23 & \textbf{0.70} & 0.65 & \underline0.67 \\
%     F1 Score $\uparrow$ & 0.32 & \textbf{0.72} & 0.67 & \underline0.70 \\ \bottomrule
%     \end{tabular}}
%     \vspace{-3ex}
% \end{table}

\begin{table}[]
    \caption{Comparative results on human motion generation from object movement task. Bold indicates best and underline indicates second best.}
    \label{tab: object2motion}
\setlength{\tabcolsep}{0.2ex}
\resizebox{\linewidth}{!}{
\begin{tabular}{lcccccccc}
\hline
Method & Hand JPE $\downarrow$ & MPJPE $\downarrow$ & MPVPE $\downarrow$ & $\mathbf{T}_{\text{root}} \downarrow$ & $\mathbf{O}_{\text{root}} \downarrow$ & $\mathbf{C}_{\text{prec}} \uparrow$ & $\mathbf{C}_{\text{rec}} \uparrow$ & F1 Score $\uparrow$ \\
\hline
GOAL~\citep{taheri2022goal} & 49.90 & 15.64 & 21.82 & 34.35 & 0.76 & \textbf{0.83} & 0.23 & 0.32 \\
OMOMO~\citep{li2023object} & \underline{24.01} & \underline{12.42} & \underline{16.67} & 18.44 & \underline{0.50} & 0.82 & \textbf{0.70} & \textbf{0.72} \\
CHOIS~\citep{li2024controllable} & - & 15.82 & - & 24.75 & - & 0.77 & 0.65 & 0.67 \\
\textbf{TCM (ours)} & \textbf{23.40} & \textbf{11.34} & \textbf{15.18} & \textbf{17.46} & \textbf{0.45} & \underline{0.82} & \underline{0.67} & \underline{0.70} \\
\hline
\end{tabular}}
    \vspace{-3ex}
\end{table}

\subsubsection{Human Motion Estimation from Egocentric and Music}
In practice, the temporal condition can be single modality (e.g., music-to-dance) or multiple modalities. In this experiment, we explore how our TCM performs when the input conditions have multiple modalities. In particular, we set up the task of human motion estimation from egocentric and music input. This task requires the alignment between the head motion with the egocentric view, while ensuring the body moves naturally with the music. We use a diffusion model that adopts the same architecture as the proposed TCM. For feature extraction, we use Jukebox~\citep{dhariwal2020jukebox} to obtain music embeddings and ResNet-50~\citep{he2016deep} to extract egocentric visual embeddings. The music and visual embeddings are first aligned via temporal contrastive loss~\cite{han2022temporal}, then fused through an MLP to produce a unified temporal condition embedding. 
% See details in Supplementary Material.

%In literature, many efforts have focused on predicting human dance motion using either egocentric video~\citep{li2023ego} or music as input~\citep{li2024lodge,tseng2023edge}. However, the task of jointly estimating human motion from both egocentric video and music remains largely unexplored. Estimating dance motion from a first-person view can help overcome challenges common in third-person settings, such as depth ambiguity and occlusion. Although this task opens the door to many exciting applications~\citep{guo2022dancevis,kim2023performing}, it is also quite challenging: the generated motion needs to keep the head movement consistent with the egocentric view, while ensuring the body moves naturally with the music. To address this challenge, we first train a diffusion model conditioned on a fused embedding of music and egocentric images. The diffusion model adopts the same architecture as the proposed Temporally Condition Mamba. For feature extraction, we use Jukebox~\citep{dhariwal2020jukebox} to obtain music embeddings and ResNet-50~\citep{he2016deep} to extract egocentric visual embeddings. These embeddings are aligned through a contrastive loss and subsequently fused via an MLP layer to produce a comprehensive condition embedding. During the sampling stage, we apply a guidance sampling technique~\citep{huang2023diffusion} to steer the generated head motion to match the estimated trajectory, which is predicted using pretrained DROID-SLAM~\citep{teed2021droid}.

\textbf{Baselines.} We compare our TCM method with egocentric motion estimation methods (PoseReg~\citep{yuan2019ego}, Kinpoly~\citep{luo2021dynamics}, EgoEgo~\citep{li2023ego}) and music-driven motion generation methods (FACT~\cite{li2021ai}, Bailando~\citep{siyao2022bailando}, EDGE~\citep{tseng2023edge}). Since our task leverages both egocentric video and music, we ensure fair comparison by equipping egocentric baselines with the Jukebox~\citep{dhariwal2020jukebox} encoder for audio processing. In addition to these baselines, we compare our work with EMM~\cite{nguyen2025egomusic}, a multimodal Mamba-based model for human motion generation. For music-based models, we inject visual features from the egocentric video using~\cite{he2016deep}. More implementation details will be released with our source code.

\textbf{Dataset and metrics.} 
% While several datasets have been proposed for single input (either egocentric or music) human motion estimation, datasets that combine egocentric and music for dance pose are still limited. We craft a dataset, called EgoAIST++ that integrates egocentric views and music specifically for human dance pose estimation.
We use the EgoExo4D dataset~\cite{grauman2024ego}, which contains approximately two hours of dance sequences, along with egocentric videos and music.
Following~\cite{li2023ego}, we evaluate our method using five standard metrics commonly used in human pose estimation task: $\mathbf{O_{\text{head}}}$, $\mathbf{T_{\text{head}}}$, MPJPE, Accel, FS. 
% To evaluate how well the generated dance motions align with the music and egocentric video, we propose a new metric called the Beat-Vision Align Score (BVAS) (see Supplementary Material).

\textbf{Results.} We report the quantitative performance of our method and other baselines in Table~\ref{tab: egomusic2motion}. The results indicate that our approach generates more physically plausible motions and achieves superior alignment with both the egocentric video and the accompanying music. For qualitative results please refer to our demo video.

% \vspace{-1.4ex}
\subsubsection{Human Motion Synthesis from Object Trajectories}
In this task, we aim to generate full-body human motion from a sequence of object trajectories~\cite{li2023object}. We adopt a two-stage approach based on a conditional diffusion framework, following the methodology proposed in~\cite{li2023object}. In the first stage, the model predicts the positions of the hands in contact with the object based on its geometry. The second stage then generates full-body poses conditioned on the predicted hand joint positions. Our TCM is used as a backbone in both stages. Specifically, the sequence of object meshes in stage one and the sequence of predicted hands position in stage two are fed into TCM as temporal conditions.

%Understanding and generating human motion within rich, contextual environments is crucial for advancing fields such as character animation, virtual and augmented reality, and robotics. 

\textbf{Dataset and metrics.} We use the Omomo dataset~\cite{li2023object}, a motion capture dataset comprising approximately 10 hours of motion data collected from 17 subjects. For consistency with prior work, we adopt the same data partitioning strategy for training and evaluation, particularly 15 subjects for training and 2 subjects for testing. We evaluate the results using metrics in~\cite{li2023object}. %The first perspective measures the difference between the generated motion and the ground truth, while the second assesses physical plausibility through contact correctness, object penetration, and foot sliding.

\textbf{Results.} We compare our method with two transformer-based baselines: GOAL~\cite{taheri2022goal} and OMOMO~\cite{li2023object}. 
We present the qualitative comparison results in Fig.~\ref{fig:object2motion}. Our proposed method generates more plausible human motions compared to OMOMO. 
% Qualitative results in Figure~\ref{fig:object2motion} show that OMOMO sometimes generates less accurate interactions, such as carrying a box instead of a monitor (top row), while our method produces more plausible motions with better object contact (bottom row). 
Quantitative results in Table~\ref{tab: object2motion} further confirm that our approach outperforms existing baselines in generating realistic object-conditioned motion.
\vspace{-1.6ex}
\subsection{Discussion}\label{sec:discussion}
\subsubsection{Limitations.} Although our proposed approach adapts well to many tasks, it has certain limitations. The Temporally Conditioned Mamba is specifically designed for time-dependent conditions, and may not be optimal for static inputs such as text~\cite{karunratanakul2023guided} or scene descriptions~\cite{huang2023diffusion}, where the condition does not share the same temporal resolution or length as the motion sequence. 
% Second, when multiple condition modalities (e.g., egocentric video and music) are present, our approach fuses them into a single representation before injection. This may obscure modality-specific temporal dynamics. Inspired by ~\cite{li2024coupled}, a promising direction for future work is to inject each modality into parallel Mamba blocks, allowing the model to process and integrate them in a more structured and disentangled manner. 
Second, in the egocentric2motion task, extreme or abrupt head movements (e.g., sudden jerks or whiplash-like motions) remain challenging and may result in less accurate motion generation. We consider this an important direction for future work.
Finally, the generated motion may still exhibit visible foot-skating and jitter artifacts. Incorporating additional kinematic constraints (e.g., velocity and acceleration losses and adding modules for explicit foot-contact modeling will likely improve motion stability and reduce such artifacts~\cite{li2024lodge}.

\subsubsection{Broader Impact.} We believe that our work presents a significant step forward in human motion learning. Our method has broad potential across areas such as animation, AR/VR, and human-robot interaction, where the generated motion must respond to dynamic inputs. While we focus on human motion learning, the core idea of TCM can be applied to other time-based tasks, such as speech~\cite{mehrish2023review, zhang2025rethinking} and physiological signal modeling~\cite{zou2025rhythmmamba, zou2024rhythmmamba}. We hope this work inspires further research into learning from complex, real-world temporal data, especially as temporally conditioned generation becomes increasingly relevant across applications.

% \vspace{-3ex}
\section{Conclusion} 
In this paper, we propose a new diffusion-based framework for human motion learning, built on the Mamba architecture and tailored for temporal conditioning. Our Temporally Conditioned Mamba (TCM) recursively integrates the conditional signal and motion sequence, enabling stronger temporal alignment and more coherent motion generation. Extensive experiments show that our method consistently outperforms other baselines, achieving state-of-the-art results across four different tasks in both motion synthesis and estimation settings. Our code and trained models will be released publicly to encourage further research.

%%%%%%%%%%%%%%%%%%%%%%%%%%%%%%%%%%%%%%%%%%%%%%%%%%%%%%%%%%%%
\bibliographystyle{ACM-Reference-Format}
% \bibliography{sample-bibliography}
\bibliography{ref_acm}
\end{document}